\documentclass[11pt]{article}

\usepackage[final]{acl}

\usepackage{times}
\usepackage{latexsym}

\usepackage{tabularx}
\usepackage{amsmath,amssymb}
\usepackage{multirow}
\usepackage{booktabs}
\usepackage{adjustbox}
\usepackage{makecell} 
\usepackage{listings}

\usepackage{colortbl}
\usepackage{xcolor}
\usepackage{tcolorbox}
\usepackage{array}
\usepackage{soul} 

\definecolor{lightblue}{RGB}{220,235,255}
\definecolor{lightyellow}{RGB}{255,245,200}

\newcommand{\posi}[1]{\colorbox{lightyellow}{#1}} 
\newcommand{\nega}[1]{\colorbox{lightblue}{#1}}

\renewcommand{\arraystretch}{1.2}
\usepackage[T1]{fontenc}

\usepackage[utf8]{inputenc}

\usepackage{microtype}

\usepackage{inconsolata}

\usepackage{graphicx}

\title{A Unified Framework to Elicit Structured Feedback \\ for Interpretable Multi-Trait Essay Scoring}

\author{
Shihang Yang\textsuperscript{1,2} \quad
Sanwoo Lee\textsuperscript{1,2} \quad
Ningning Zhao\textsuperscript{3} \quad
Yunfang Wu\textsuperscript{1,2}\thanks{Corresponding author.} \\
\textsuperscript{1}National Key Laboratory for Multimedia Information Processing, Peking University \\
\textsuperscript{2}School of Computer Science, Peking University \\
\textsuperscript{3}School of Chinese Language and Literature, Beijing Normal University \\
\texttt{sh\_yang@stu.pku.edu.cn},
~~~\texttt{wuyf@pku.edu.cn}
}

\begin{document}
\maketitle
\begin{abstract}
Multi-trait Automated Essay Scoring (AES) requires rubric-grounded reasoning across interdependent traits, rather than isolated score prediction.
Existing feedback-enhanced methods often decouple feedback from scoring or assess traits independently, weakening score--feedback consistency and rubric alignment.
We propose \textbf{HiFTS}, a unified autoregressive framework that generates hierarchical CoT feedback before predicting trait-level and holistic scores.
HiFTS distills rubric-grounded hierarchical CoT feedback from a teacher LLM and trains student models to jointly generate feedback and scores. 
HiFTS further applies Group Relative Policy Optimization with a composite reward balancing score agreement, calibration, feedback quality, and structural validity.
At inference, a lightweight global prior provides holistic guidance to reduce drift during long-form reasoning.
We also introduce \textbf{CFMS-34}, a Chinese multi-trait AES dataset with 951 essays annotated with holistic scores and 34 rubric-based traits. 
Experiments on CFMS-34 and ASAP++ show that HiFTS achieves strong holistic and trait-level scoring while producing coherent, rubric-aligned feedback. 
\textit{We release our data and code at} \url{https://github.com/Atiyahsama/HiFTS}.


\end{abstract}

\section{Introduction}

Grounding writing assessment in multi-faceted feedback is essential for providing learners with constructive guidance and ensuring transparency of automated scoring. Traditionally, however, automated essay scoring (AES) models have focused on score-only predictions, pursuing high agreement with human raters at the cost of leaving the underlying decision process black-box to the end users \citep{taghipour-ng-2016-neural, yang-etal-2020-enhancing, wang-liu-2025-mes}. While human-crafted features add partial interpretability via linear models \citep{yupei-renfen-2021-prompt} or post-hoc activation alignment \citep{fiacco-etal-2023-towards}, such model-centric explanations remain challenging for humans to interpret.

Driven by breakthroughs in large language models (LLMs) \citep{ouyang2022training, achiam2023gpt}, recent AES studies have sought to mirror human-like evaluation by predicting scores alongside natural language rationales. While early attempts harnessed LLMs' inherent reasoning potential via in-context learning \citep{yancey2023rating, stahl-etal-2024-exploring} and sophisticated prompting \citep{wang2025autoscore}, they found a substantial gap in scoring accuracy compared to traditional supervised models \citep{lee-etal-2024-unleashing}. Consequently, studies have employed supervised fine-tuning (SFT) or reinforcement learning (RL), which demonstrate a narrowed accuracy gap while preserving LLMs' strengths in generating rationales \citep{chu-etal-2025-rationale, li-pan-2025-ceaes, do2025radme}. Notably, these methods are increasingly evaluated on \textit{multi-trait} AES tasks \cite{chu-etal-2025-rationale, do2024samrl} to test their capacity for fine-grained evaluation across traits with intricate dependencies.


Despite the improvements, current supervised approaches lack a unified framework that generates multi-trait feedback and scores jointly in a single autoregressive process. For instance, feedback is generated from an external LLM to augment inputs for training score-only model \citep{chu-etal-2025-rationale}. Methods that natively generate feedback exhibit either a lack of inter-trait dependencies due to independent trait-by-trait generation \cite{do2025radme}, or a disconnect between the score and the rationale stemming from a separate regression head \cite{li-pan-2025-ceaes}.

Motivated by these insights, we propose \textbf{HiFTS} (Hierarchical Feedback-to-Score Reasoning for Multi-Trait Essay Scoring), a unified framework that reinforces LLMs to elicit structured multi-trait feedback and scores. Specifically, our hierarchical input template prompts the model to capture inter-trait relations by generating global-to-local rationales followed by rubric-aligned scores. After a warm-up SFT with CoT traces prompted by the template, we use group-relative policy optimization (GRPO) \citep{deepseekmath2024} to align the feedback traces with ground-truth trait scores as verifiable rewards. At inference time, the generation process is further conditioned on a prior score predicted by a lightweight BERT model \citep{devlin-etal-2019-bert}, which improves scoring stability and accuracy.




Furthermore, we present \textbf{CFMS-34}, a Chinese multi-trait AES dataset with 34 fine-grained trait scores annotated by human experts. This dataset addresses a major gap where recent multi-trait AES methods are predominantly validated and optimized on English benchmarks \citep{ridley2021multitrait, do2024arts, wang-liu-2025-mes}. CFMS-34 serves as a rigorous testbed for cross-lingual evaluation, and we find that recent high-performing methods on the standard English benchmark, ASAP++ \citep{mathias2018asappp}, do not necessarily succeed on CFMS-34. In contrast, experiments on both CFMS-34 and ASAP++ demonstrate that our proposed HiFTS improves both LLMs' scoring accuracy and the quality of feedback. Fine-grained and ablation analyses further show that these gains come from hierarchical CoT supervision, GRPO-based alignment, and essay-specific prior guidance at inference.


Our contributions are summarized as follows:
\begin{itemize}
    \item We construct and release \textbf{CFMS-34}, a fine-grained 
     multi-trait Chinese AES dataset curated by expert annotators, promoting cross-lingual validation of multi-trait AES methods.
    \item We propose \textbf{HiFTS}, a unified LLM post-training framework that elicits structured feedback and multi-trait knowledge.
    \item  During inference, we coordinate the LLM with a lightweight BERT model to enhance scoring accuracy and stability via prior-guilded generation.
\end{itemize}

\begin{table*}[t]
    \centering
    \tiny
    {
    \setlength{\aboverulesep}{0.3ex}
    \setlength{\belowrulesep}{0.3ex}
    \resizebox{\textwidth}{!}{
    \begin{tabular}{cll}
        \toprule
        \textbf{Dimension} & \textbf{Code} & \textbf{Definition / Trait Description} \\
        \midrule
        \textbf{Content} & C01--C03 & Authenticity, sufficiency, and novelty of selected materials. \\
                         & C04--C05 & Alignment with the central theme; depth of thought and originality. \\
                         & C06 & Multi-perspective thinking and appropriateness of material presentation. \\
        \midrule
        \textbf{Structure} & S01 & Logical paragraph division and structural clarity. \\
                           & S02--S03 & Use of foreshadowing, transitions, and cohesive devices. \\
                           & S04--S05 & Structural coherence (e.g., ``General-Specific-General'') and thread development. \\
                           & S06--S08 & Relevance among paragraphs, sentences, and the central theme. \\
                           & S09--S10 & Balance between detailed and concise writing (emphasis on key points). \\
                           & S11--S12 & Quality of the opening (engagement) and ending (summary/insight). \\
        \midrule
        \textbf{Expression} & E01--E02 & Sentence fluency, grammatical correctness, and sophistication of word choice. \\
                            & E03--E04 & Degree of descriptive detail and logical narrative order. \\
                            & E05--E07 & Use of rhetorical devices, varied expression modes, and writing techniques. \\
                            & E08--E10 & Observational perspective (holistic vs.\ focused) and narrative methods. \\
                            & E11--E12 & Precision in capturing key characteristics of characters/events with clear prioritization. \\
        \midrule
        \textbf{Conventions} & Cv01--Cv02 & Standard formatting and accurate punctuation usage. \\
                           & Cv03--Cv04 & Orthographic correctness and aesthetic handwriting. \\
        \bottomrule
    \end{tabular}}
    }
    \caption{Definition of scoring traits in the CFMS-34 dataset.}
    \label{tab:trait_definitions}
\end{table*}

\section{Related Work}

\paragraph{Multi-trait essay scoring.}
Automated essay scoring (AES) has progressed from feature- and semantics-based scoring systems to neural and pre-trained language model based approaches 
\citep{foltz1999intelligent,attali-burstein-2006,taghipour-ng-2016-neural,yang-etal-2020-enhancing,wang-etal-2022-use,das-etal-2024-transformer}. 
Multi-trait AES extends holistic scoring by predicting scores for rubric-defined dimensions, enabling more fine-grained assessment of writing quality 
\citep{mathias2018asappp,mathias2020traits,ridley2021multitrait,kumar2021manyhands}. 
Recent studies improve multi-trait scoring with cross-prompt modeling, trait-aware representations, rubric-assisted features, mixture-of-experts, and graph-based trait interactions 
\citep{do2023prompttrait,wang-liu-2025-mes,eltanbouly-etal-2025-trates,li-ng-2025-graph,zhaoMultiknowledgeEnhancedGraph2026}. 
Another line of work casts multi-trait AES as autoregressive score generation, where language models generate structured trait scores and can be further optimized with score-related objectives 
\citep{do2024arts,do2024samrl}. 
However, most approaches model trait dependencies implicitly, with shallow relation modules, or through order-sensitive score generation 
\citep{ridley2021multitrait,do2023prompttrait,do2024arts,li-ng-2025-graph}. 
Concurrent work studies trait-aware post-training for score-only autoregressive generation \citep{wang-etal-2026-tapo}; in contrast, HiFTS integrates hierarchical CoT feedback with score generation, so that trait-level scores are produced after global-to-local rubric-grounded reasoning.

\paragraph{Feedback-based scoring and alignment.}
Recent AES work explores LLMs for rubric-aware scoring and feedback generation through prompting, supervised training, and distillation 
\citep{yancey2023rating,stahl-etal-2024-exploring,lee-etal-2024-unleashing,wang2025autoscore,harada-etal-2026-automated}. 
To improve reliability, subsequent methods incorporate LLM-generated rationales or feedback into supervised scoring pipelines, jointly train scoring and feedback modules, or distill score-guided feedback from stronger teacher models 
\citep{chu-etal-2025-rationale,li-pan-2025-ceaes,do2025radme}. 
Despite their effectiveness, these methods often separate reasoning from scoring: feedback may be produced externally, numerical scores may be predicted by separate heads, or trait-level feedback may be generated independently. 
Post-training methods such as RLHF, preference optimization, and GRPO align generative models with task-specific, verifiable, or score-related rewards 
\citep{schulmanProximalPolicyOptimization2017,ouyang2022training,rafailovDirectPreferenceOptimization2023,deepseekmath2024,do2024samrl}. 
Our work follows this direction but aligns a single autoregressive process that produces structured multi-trait feedback then generates calibrated trait and holistic scores.

\section{CFMS-34 Dataset Construction}
\label{sec:cfms34}

\subsection{Data Collection and Curation}
In this paper, we present \textbf{CFMS-34}, a Chinese multi-trait essay scoring dataset featuring high-level dimensions that encompass fine-grained sub-traits, constructed through a rigorous pedagogical assessment process. The dataset consists of 951 essays written by primary school students during in-class examinations. To ensure consistent writing conditions and reflect authentic classroom settings, all essays were completed in response to unit-level writing prompts from textbooks within a 50-minute time limit. 

\begin{figure}[ht]
\centering
\includegraphics[width=0.9\linewidth]{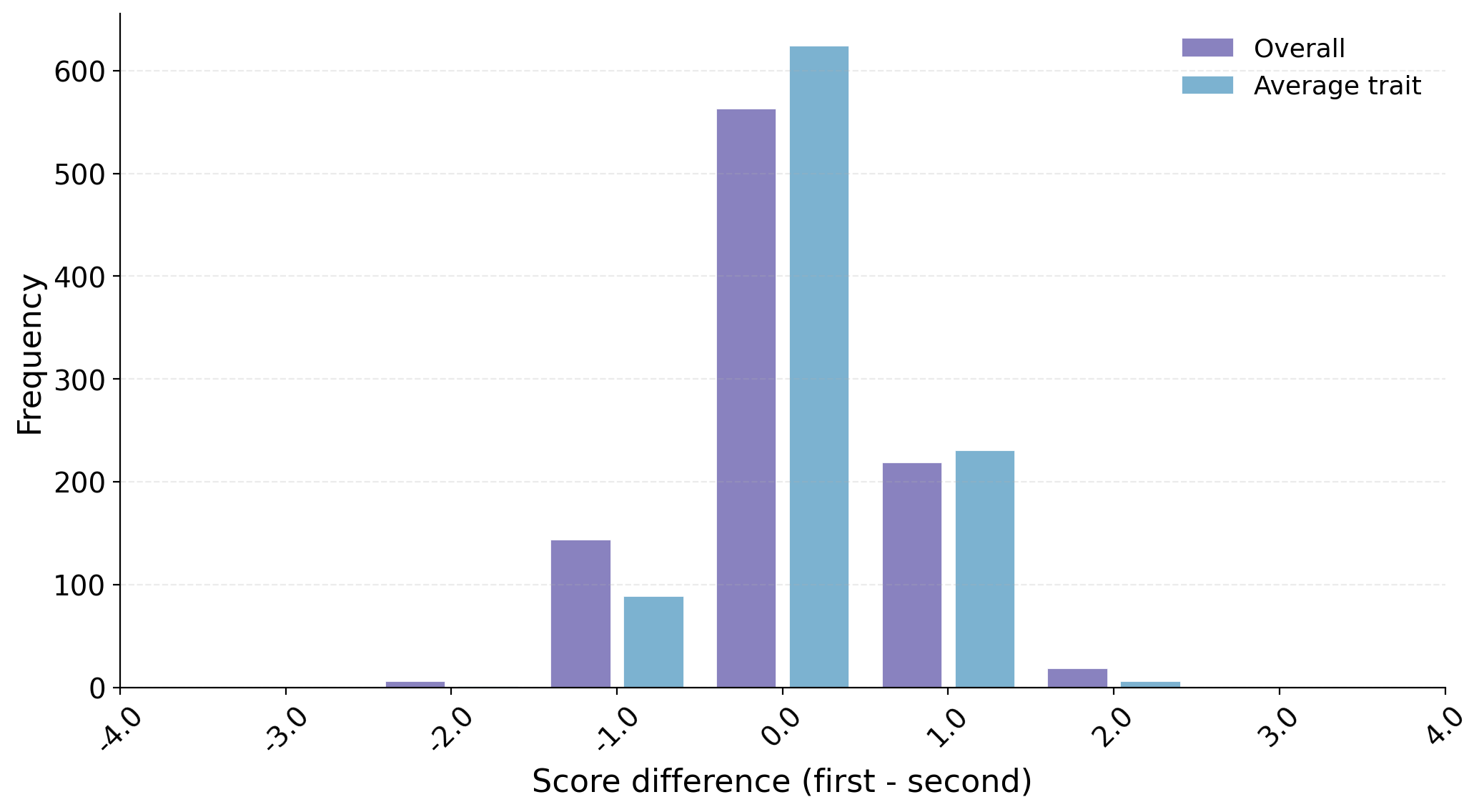}
\caption{Distribution of score differences between the two raters.}
\label{fig:score_diff}
\end{figure}

Two trained experts in Chinese language education independently scored each essay based on strict guidelines. These guidelines define 34 fine-grained sub-traits categorized into four high-level dimensions: Content, Structure, Expression, and Conventions, as summarized in Table~\ref{tab:trait_definitions}. Ultimately, the annotation process provides 34 sub-trait scores and an overall score, all on a scale of 0 to 5. For 97.4\% of the samples, the experts agreed on the overall score within one-point difference, and the human--human QWK on the overall score is 0.57, indicating a solid inter-rater consistency, as shown in Figure~\ref{fig:score_diff}. Across the 34 sub-traits, the average agreement is 0.48 exact match, 0.90 within one point, and 0.41 QWK (see Appendix~\ref{app:trait_irr} for the full per-trait table).

\begin{table}[ht]
    \centering
    \small
    \begin{tabular}{ll}
        \toprule
        \textbf{Statistic} & \textbf{Value} \\
        \midrule
        Total Essays & 951 \\
        Language & Chinese (Simplified) \\
        Raters per Essay & 2 (Human Experts) \\
        Avg.\ Essay Length & 492.4 Chinese characters \\
        Score Dimensions & Overall + 34 Sub-traits \\
        Score Range & 0.0 -- 5.0 \\
        Overall Score (mean $\pm$ std) & $3.64 \pm 0.74$ \\
        Human--Human QWK (overall) & 0.57 \\
        \bottomrule
    \end{tabular}
    \caption{Statistics of the CFMS-34 dataset.}
    \label{tab:dataset_stats}
\end{table}

We split CFMS-34 into training, development, and test sets using an 8:1:1 ratio. To ensure a reliable evaluation benchmark, the test set explicitly reserves samples with perfect inter-rater agreement on the overall score; this design is intended to reduce label noise in the test labels, rather than to simplify the task. The dataset statistics are shown in Table~\ref{tab:dataset_stats} for reference.

\begin{table*}[t]
\centering
\small
\begin{tabular}{lccccc}
\toprule
\textbf{Dataset} 
& Language
& Expert Annotation 
& Dual Raters 
& Multi-trait 
& Fine-grained \\
\midrule
EFCAMDAT \cite{geertzen2013automatic} & EN & $\times$ & $\times$ & $\times$ & $\times$ \\
ICLE \cite{granger2009icle} & EN & $\checkmark$ & $\times$ & $\checkmark$ & $\times$ \\
TOEFL11 \cite{blanchard2013toefl11} & EN & $\checkmark$ & $\times$ & $\times$ & $\times$ \\
FCE \cite{yannakoudakis2011new} & EN & $\checkmark$ & $\checkmark$ & $\times$ & $\times$ \\
ASAP++ \cite{mathias2018asappp} & EN & $\checkmark$ & $\checkmark$ & $\checkmark$ & $\checkmark$ \\
TOREE \cite{zhuang2024toree} & CN & $\checkmark$ & $\times$ & $\checkmark$ & $\times$ \\
CEDCC \cite{wu2023cedcc} & CN & $\checkmark$ & $\checkmark$ & $\checkmark$ & $\times$ \\
\midrule
CFMS-34 (ours) & CN & $\checkmark$ & $\checkmark$ & $\checkmark$ & $\checkmark$ \\
\bottomrule
\end{tabular}
\caption{Comparison with existing essay scoring datasets.}
\label{tab:dataset_comparison}
\end{table*}







\subsection{Comparison with Existing Datasets}

Table~\ref{tab:dataset_comparison} compares CFMS-34 with representative English and Chinese essay scoring datasets.
Like several prior datasets, CFMS-34 provides expert annotations, dual-rater scoring, and multi-trait labels.
Its main distinction lies in the fine-grained rubric, which defines detailed sub-traits tailored to Chinese essay writing.
This design enables more precise evaluation of writing quality across multiple dimensions and makes CFMS-34 a useful benchmark for multi-dimensional essay scoring. 

\subsection{Benchmark Performance of Frontier LLMs}

To examine the difficulty of CFMS-34, we evaluate recent large language models on the overall scoring task, including GPT, Gemini, Doubao, Qwen, and DeepSeek series.
We report quadratic weighted kappa (QWK) against human scores.
The frontier-LLM baselines use the same zero-shot scoring prompt as Appendix~\ref{sec:appendix_system_prompt}. We generate one score per sample using each provider's default decoding settings and extract the numeric value from the \texttt{[Overall Score]} field.

\begin{table}[ht]
    \centering
    \small
    {
    \setlength{\aboverulesep}{0.5ex}
    \setlength{\belowrulesep}{0.5ex}
    \renewcommand{\arraystretch}{1.0}
    \setlength{\tabcolsep}{3pt}
    \begin{tabular}{lcc}
        \toprule
        \textbf{Model} & \textbf{Mean Score} & \textbf{QWK} \\
        \midrule
        GPT-4.1 \cite{openai2025gpt41} & 3.3475 & 0.4222 \\
        Gemini-3 \cite{google_gemini3} & 3.1937 & 0.4363 \\
        Doubao-Seed-1.6 \cite{doubao_seed16} & 2.0743 & 0.1595 \\
        Qwen3-max \cite{qwen3max2025} & 3.4059 & 0.4201 \\
        DeepSeek-V3 \cite{deepseekv3} & 3.5792 & 0.4130 \\
        DeepSeek-R1 \cite{deepseek2025r1} & 3.0396 & 0.3271 \\
        \midrule
        \textbf{Human (avg.)} & 3.64 & 0.57 \\
        \bottomrule
    \end{tabular}
    }
    \caption{LLM performance on CFMS-34.}
    \label{tab:llm_benchmark}
\end{table}

\begin{figure*}[t]
    \centering
    \includegraphics[width=\textwidth]{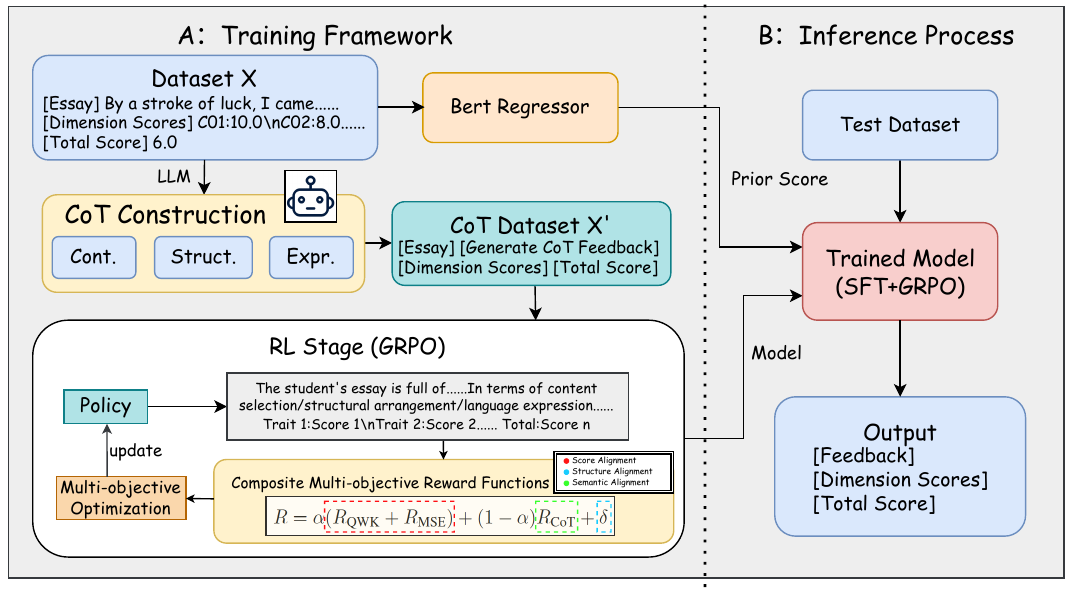}
    \caption{Overview of HiFTS. HiFTS first constructs hierarchical CoT feedback with an LLM teacher, then trains the student model with SFT and GRPO. During inference, a BERT-based global prior guides feedback generation and score prediction for more consistent multi-trait AES.}
    \label{fig:overview}
\end{figure*}

As shown in Table~\ref{tab:llm_benchmark}, current LLMs achieve only moderate agreement with human annotations, with the best QWK below 0.44.
The Human QWK of 0.57 measures agreement between the two expert raters, whereas model QWKs are computed against the resolved human scores; the two quantities are reported together for reference but are not strictly comparable.
Their mean scores also reveal calibration issues, such as substantial under-prediction by Doubao-Seed-1.6 and slight over-prediction by several other models.
These results suggest that CFMS-34 remains challenging for current LLMs and motivates structured reasoning methods \mbox{beyond} flat score prediction.

\section{Method}
As shown in Fig.~\ref{fig:overview}, \textbf{HiFTS} formulates multi-trait AES as a unified autoregressive process that generates rubric-aware hierarchical feedback before trait-level and holistic scores.
Training first uses teacher-generated CoT supervision for SFT (Section~\ref{sec:warmup_sft}) and then applies GRPO with a composite reward (Section~\ref{sec:rl_reward}); inference further conditions decoding on a BERT-based global prior to improve score consistency (Section~\ref{sec:global_prior}).

\subsection{Warmup SFT for Hierarchical Feedback-to-Score Generation}
\label{sec:warmup_sft}

Directly applying RL with only final score-level rewards may fail to incentivize the model to faithfully capture cross-trait dependencies within its reasoning traces. To address this, we introduce a warmup SFT stage prior to RL, designed to internalize the capacity for structured hierarchical reasoning. Using Gemini-3 \cite{google_gemini3} as a teacher model, we construct CoT feedback following a top-down hierarchy: \textbf{global understanding of essay} $\rightarrow$ \textbf{analysis of a dimension} $\rightarrow$ \textbf{sub-trait evaluation}. To assure alignment of the generated feedback with the rubric guidelines and labeled scores, prompting the teacher model is explicitly conditioned on the essay, the rubric definitions, and the labeled scores, see Appendix~\ref{sec:appendix_prompt} for full template.

Subsequently, the generated feedback is appended with the labeled trait-level and overall scores to construct a unified feedback-to-score target sequence. We fine-tune the student model on these sequences using an instructional prompt that can frame the target feedback-to-score sequence as its natural continuation. The corresponding template is provided in Appendix~\ref{sec:appendix_system_prompt}. To be concrete, the target sequence is the concatenation of the feedback and the labeled scores:
\begin{equation}
Y = [Y_{\text{CoT}} ; Y_{\text{Score}}],
\end{equation}
and we use the standard token-level cross-entropy loss during SFT:
\begin{equation}
\mathcal{L}_{\text{SFT}}
=
-\sum_{t=1}^{T}
\log P_{\theta}(y_t \mid y_{<t}, X),
\end{equation}
where $X$ is the input essay and $y_t$ is the target token at step $t$. To summarize, this warmup SFT stage ensures the student's CoT trace tightly anchors its eventual scoring decisions, serving as an initialization point for subsequent RL stage.

\subsection{Boosting Feedback Quality and Scoring Accuracy via Reinforcement Learning}
\label{sec:rl_reward}

To jointly optimize scoring precision, feedback quality, and formatting compliance, we define a comprehensive multi-objective reward function $R$, formulated as:
\begin{equation}
R = \alpha (R_{\text{QWK}} + R_{\text{MSE}}) + (1-\alpha) R_{\text{CoT}} + \delta,
\end{equation}
where $\alpha \in [0, 1]$ serves as a balancing coefficient that controls the trade-off between score-oriented alignment and feedback-oriented alignment. The structural term $\delta$ is a fixed rule-based bonus or penalty and does not introduce additional \mbox{hyperparameters} beyond $\alpha$.


$R_{\text{QWK}}$ captures both overall- and trait-level agreement with human annotations. We compute holistic QWK over a short sliding window of recent generations to stabilize the reward, and compute trait-level QWK over the traits:
\begin{equation}
R_{\text{QWK}}
=
\kappa(\hat{\mathbf{s}}_{t-h+1:t}, \mathbf{s}_{t-h+1:t})
+
\kappa(\hat{\mathbf{z}}^{(m)}_{t}, \mathbf{z}^{(m)}_{t})
\end{equation}
where $\kappa$ denotes QWK, $h=64$ is the sliding-window size over the most recent generations, and $\hat{\mathbf{z}}^{(m)}_{t}$ and $\mathbf{z}^{(m)}_{t}$ denote the predicted and gold scores over the $m$ traits.

$R_{\text{MSE}}$ is defined as the negative mean squared error between the predicted and gold scores, thereby penalizing larger numerical deviations and encouraging accurate score prediction.

$R_{\text{CoT}}$ measures semantic alignment between generated feedback and teacher CoT references.
We compute this reward as the cosine similarity between sentence embeddings of generated and teacher feedback:
\begin{equation}
R_{\text{CoT}}
=
\cos \left( f(\hat{c}_t), f(c_t) \right)
=
\frac{
f(\hat{c}_t)^\top f(c_t)
}{
\|f(\hat{c}_t)\| \, \|f(c_t)\|
},
\end{equation}
where $\hat{c}_t$ and $c_t$ denote the generated and teacher feedback, respectively, and $f(\cdot)$ is instantiated as \texttt{bge-small-zh-v1.5} for CFMS-34 and \texttt{all-MiniLM-L6-v2} for ASAP++.


This formulation balances scoring accuracy and feedback quality without introducing multiple reward weights.
The value of $\alpha$ is selected on the development set, as discussed in Section~\ref{sec:hyperparameter_analysis}.

We adopt Group Relative Policy Optimization (GRPO) for reinforcement learning. 
For each prompt, a group of candidate responses is sampled and evaluated with the reward function above, and the policy is updated according to their group-relative rewards.

\subsection{Robust Inference via Global Prior Guidance}
\label{sec:global_prior}

Long-form hierarchical CoT generation can suffer from semantic drift, where the model gradually departs from its initial assessment as the reasoning chain unfolds.
This is particularly problematic for multi-trait AES, since trait-level judgments should remain consistent with the overall writing quality.

To mitigate this issue, we introduce a lightweight global prior at inference time.
We use a BERT-based regressor to estimate holistic essay quality.
Given an essay $X$, the regressor produces a coarse prior score $S_{\text{prior}}$, which is inserted into the system prompt before decoding with the GRPO-aligned LLM as guidance.

Importantly, $S_{\text{prior}}$ is not used as the final prediction.
Instead, it serves as a soft semantic anchor that guides the model toward a plausible scoring region, while still allowing the LLM to generate rubric-grounded feedback and final scores autoregressively.
This strategy introduces no additional decoding-time hyperparameters.

Formally, the generation process is conditioned on both the essay and the prior score:
\begin{equation}
\begin{aligned}
P(Y_C,\!S\mid X,\!S_{\text{prior}})
&=P(Y_C\mid X,\!S_{\text{prior}}) \\
&\cdot P(S\mid Y_C,\!X,\!S_{\text{prior}}),
\end{aligned}
\end{equation}
where $Y_C$ denotes the generated hierarchical feedback and $S$ denotes the predicted trait-level and holistic scores.

By anchoring generation with an initial estimate of holistic quality, prior-guided decoding reduces inconsistencies between local trait analyses and final scoring decisions.

\begin{table*}[ht]
    \small
    \centering
    \setlength{\tabcolsep}{8pt}
    \begin{tabular}{l|c|cccc|ccc}
        \toprule
        \multirow{2}{*}{\textbf{Method}} & \multirow{2}{*}{\textbf{Explain.}}
        & \multicolumn{4}{c|}{\textbf{CFMS-34}} & \multicolumn{3}{c}{\textbf{ASAP++}} \\[-0.5mm]
        & & Overall & Traits & MSE & WinRate & Prompts & Traits & WinRate \\
        \midrule
        HISK & $\times$ & 0.405 & 0.309 & 1.374 & - & 0.644 & 0.611 & - \\
        STL-LSTM & $\times$ & 0.484 & 0.365 & 1.454 & - & 0.684 & 0.656 & - \\
        MTL-BiLSTM & $\times$ & 0.519 & 0.396 & 1.306 & - & 0.665  & 0.638 & - \\
        ArTS & $\times$ & 0.264 & 0.149 & 1.960 & - & 0.717 & 0.695 & - \\
        RMTS & $\checkmark$ & 0.296 & 0.303 & 1.431 & - & 0.720 & 0.704 & - \\
        \midrule
        HiFTS-SFT$_{\text{Qwen2.5}}$ & $\checkmark$ & 0.528 & 0.387 & 1.447 & Ref. & 0.712 & 0.674 & Ref. \\
        HiFTS$_{\text{Qwen2.5}}$ & $\checkmark$ & 0.656 & 0.439 & 0.957 & 0.673 & 0.718 & 0.704 & 0.550 \\
        \midrule
        HiFTS-SFT$_{\text{Qwen3}}$ & $\checkmark$ & 0.557 & 0.395 & 1.379 & Ref. & 0.720 & 0.685 & Ref. \\
        HiFTS$_{\text{Qwen3}}$ & $\checkmark$ & \textbf{0.677} & \textbf{0.453} & \textbf{0.741} & \textbf{0.706} & \textbf{0.726} & \textbf{0.711} & \textbf{0.613} \\
        \bottomrule
    \end{tabular}
    \caption{Main scoring and feedback results on CFMS-34 and ASAP++.}
    \label{tab:main_results}
\end{table*}

\section{Experimental Setup}
\label{sec:setup}

\paragraph{Datasets.}
We evaluate HiFTS on CFMS-34 and ASAP++ \citep{mathias2018asappp}, covering both Chinese and English multi-trait essay scoring.
For CFMS-34, we use the 20 core traits identified in our rubric analysis as the primary reasoning targets, with scores obtained by summing the two raters' 0--5 annotations into a 0--10 scale for training and evaluation.
The same 20-trait subset (Appendix~\ref{sec:appendix_traits20}) is applied to all methods, including baselines, while the released data retain all 34 trait annotations.
ASAP++ extends the ASAP benchmark with human-annotated trait scores for English essays across eight prompts, enabling evaluation of both holistic and trait-level scoring.

\paragraph{Evaluation Metrics.}
As shown in Table~\ref{tab:main_results}, we report quadratic weighted kappa (QWK) for scoring agreement: \textit{Overall} denotes holistic QWK on CFMS-34, \textit{Prompts} denotes the average holistic QWK over the eight ASAP++ prompts, and \textit{Traits} denotes average trait QWK.
We also report MSE on CFMS-34 for numerical calibration.
For reasoning quality, we report \textit{WinRate} from pairwise comparisons judged by DeepSeek-V3.2, used only between each SFT model and its GRPO-aligned counterpart on each dataset. 

\paragraph{Training Details.}
We fine-tune Qwen2.5-7B and Qwen3-4B with a two-stage pipeline: supervised fine-tuning on hierarchical CoT data followed by GRPO-based alignment.
For global prior guidance, we train a BERT-based regressor initialized from \texttt{bert-base-chinese} for CFMS-34 and \texttt{bert-base} for ASAP++ to estimate holistic essay quality.
During GRPO, we sample $G=4$ responses per input, set the KL coefficient to $\beta=0.02$, and use a learning rate of $1\times10^{-6}$.
For the composite reward, we set $\alpha=0.8$ based on development-set performance and use a fixed structural bonus/penalty of $\delta=0.1$.
Results are reported from one fixed-seed run, with checkpoints selected by development-set holistic QWK.

Since QWK is a set-level metric, we compute the holistic QWK reward over a sliding window of the most recent $h=64$ generations during GRPO.
At inference time, the model generates structured outputs with \texttt{[Analysis]}, \texttt{[Trait Scores]}, and \texttt{[Overall Score]} fields, conditioned on the essay, rubric prompt, and, when enabled, the global score prior.
We decode with temperature $0.6$ and \mbox{top-$p=0.9$} for all runs.

\paragraph{Baselines.}
We compare HiFTS with representative essay scoring models, including HISK \citep{cozma2018automated}, STL-LSTM \citep{dong2017attention}, MTL-BiLSTM \citep{kumar2021manyhands}, and ArTS \citep{do2024arts}.
We also include RMTS \citep{chu-etal-2025-rationale}, a rationale-enhanced multi-trait scoring model that uses trait-wise rationales generated by an external LLM.
For ASAP/ASAP++, we follow \citet{do2024samrl} and report published results for traditional baselines when the evaluation settings are \mbox{comparable} to ours.

\begin{table*}[t]
    \small
    \centering
    \setlength{\tabcolsep}{3.5pt}
    \begin{tabular}{l|ccccccccccc|c}
        \toprule
        \textbf{Method}
        & \textbf{Over.} & \textbf{Cont.} & \textbf{PA} & \textbf{Lang.} & \textbf{Nar.} & \textbf{Org.}
        & \textbf{Conv.} & \textbf{WC} & \textbf{SF} & \textbf{Style} & \textbf{Voice} & \textbf{Mean} \\
        \midrule
        HISK & 0.718 & 0.679 & 0.697 & 0.605 & 0.659 & 0.610 & 0.527 & 0.579 & 0.553 & 0.609 & 0.489 & 0.611 \\
        STL-LSTM & 0.750 & 0.707 & 0.731 & 0.640 & 0.699 & 0.649 & 0.605 & 0.621 & 0.612 & 0.659 & 0.544 & 0.656 \\
        MTL-BiLSTM & \textbf{0.764} & 0.685 & 0.701 & 0.604 & 0.668 & 0.615 & 0.560 & 0.615 & 0.598 & 0.632 & 0.582 & 0.638 \\
        ArTS & 0.751 & 0.730 & \textbf{0.751} & 0.698 & \textbf{0.725} & 0.672 & 0.668 & 0.679 & 0.678 & \textbf{0.721} & 0.570 & 0.695 \\
        \midrule
        HiFTS-PPO w/o prior & 0.751 & 0.715 & 0.730 & 0.662 & 0.708 & 0.645 & 0.640 & 0.652 & 0.625 & 0.690 & 0.565 & 0.671 \\
        HiFTS w/o prior & 0.754 & 0.722 & 0.745 & 0.685 & 0.712 & 0.670 & 0.635 & 0.668 & 0.662 & 0.688 & 0.597 & 0.685 \\
        HiFTS-PPO & 0.755 & 0.728 & 0.741 & 0.695 & \textbf{0.725} & 0.668 & 0.675 & 0.678 & 0.680 & 0.712 & 0.615 & 0.697 \\
        HiFTS & 0.758 & \textbf{0.734} & 0.749 & \textbf{0.702} & 0.724 & \textbf{0.681} & \textbf{0.684} & \textbf{0.686} & \textbf{0.688} & 0.715 & \textbf{0.626} & \textbf{0.704} \\
        \bottomrule
    \end{tabular}
    \caption{Trait-level QWK on ASAP++ with Qwen2.5. The best result for each trait is shown in bold.}
    \label{tab:asap_trait_results}
\end{table*}

\begin{table*}[t]
\centering
\small
\setlength{\tabcolsep}{6pt}
\renewcommand{\arraystretch}{1.18}

\begin{tabular}{
>{\raggedright\arraybackslash}p{0.485\textwidth}|
>{\raggedright\arraybackslash}p{0.485\textwidth}
}
\toprule
\textbf{Flat feedback (DeepSeek-Reasoner)} 
& \textbf{Hierarchical CoT feedback (Ours)} \\
\midrule

\parbox[t]{\linewidth}{

This essay revolves around the interaction between “me” and the hamster, with genuine emotions, but it is overall \nega{quite simple}, lacking in content richness and loose in structure.

The opening directly introduces the hamster’s appearance and habits, which is concise but \nega{lacks appeal}.

The main body describes playing, feeding, breeding, and changes in the relationship, but the material is \nega{organized haphazardly}, with a lack of natural transitions between paragraphs—for example, from “I loved watching it eat sunflower seeds” to “it gave birth to nine babies,” resulting in a \nega{logical leap}.

Detail description is \nega{insufficient}; for instance, the process of eating sunflower seeds is attempted but \nega{not vivid enough}.

While emotional expression is sincere, the depth of thought is \nega{lacking}, failing to explore deeper insights.

The ending \nega{suddenly confesses feelings}, appearing somewhat rushed.
The language is fluent but plain, with basically no rhetorical devices or varied expressions.

Observation follows \nega{a certain order} but lacks clear prioritization.
Overall \ldots

\vspace{2pt}
}

&

\parbox[t]{\linewidth}{

{\scshape L1 Global Essay Understanding} $\Rightarrow$

The student essay ``Little Hamster'' is \ldots.

\vspace{3pt}

{\scshape L2 Dimension Analysis}

\textbf{Content} $\Rightarrow$\\[1pt]
The essay includes \ldots These materials are \ldots (C02), \posi{and they are} \ldots theme (C04). The author \ldots \posi{in a \ldots order} (E12). \posi{In particular}, the description of \ldots is vivid and expressive (E11). The essay also elaborates \ldots engaging. \posi{However}, the essay’s \ldots lacks deeper reflection (C05), and the \posi{integration between} multi-perspective content and material selection \ldots (C06), \posi{while handling} of transitional materials appears somewhat rigid (S10).

\vspace{3pt}

\textbf{Structure} $\Rightarrow$\\[1pt]
In terms of structural organization, the essay \posi{begins with} \ldots (S11), and the conclusion remains \ldots (S12). The content of \ldots (S06), and \ldots \posi{maintain a connection} to topic sentences (S07). The preceding and subsequent content \posi{is also presented} with \ldots (S02). \posi{However}, transitions between \ldots (S03), \posi{resulting in} \ldots (S08).

\vspace{3pt}

\textbf{Expression} $\Rightarrow$ \ldots

\vspace{8pt}
}

\\
\bottomrule
\end{tabular}
\vspace{2pt}
\caption{A qualitative case study comparing flat feedback and hierarchical CoT feedback on the same essay.}
\label{tab:cot_compare}
\end{table*}

\section{Results and Analysis}
\label{sec:results}

\subsection{Scoring Performance}

Table~\ref{tab:main_results} reports results on CFMS-34 and ASAP++.
Although existing baselines remain competitive on ASAP++, their performance drops markedly on CFMS-34, suggesting that Chinese multi-dimensional essay scoring poses additional challenges.
In contrast, HiFTS performs consistently across both datasets, with GRPO further improving CFMS-34 Overall QWK, Traits QWK, and MSE, as well as ASAP++ prompt-averaged and trait-level QWK.
The Qwen3-based model with GRPO achieves the best results, reaching 0.677 Overall QWK and reducing MSE to 0.741.
These results indicate that hierarchical CoT supervision provides a strong foundation for rubric-aware scoring, while reward-based alignment improves consistency and calibration.

\subsection{Fine-grained Trait Analysis}

\begin{figure}[ht]
    \centering
    \includegraphics[width=\linewidth]{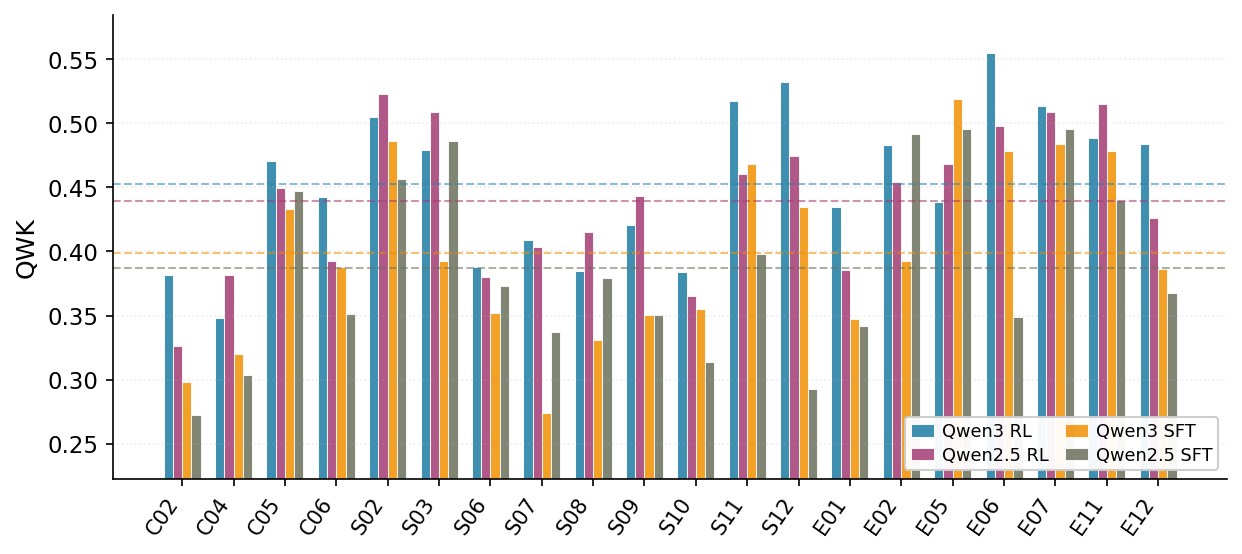}
    \caption{Trait-level QWK on CFMS-34, where dashed lines denote averages values.}
    \label{fig:cfms_trait_results}
\end{figure}

Figure~\ref{fig:cfms_trait_results} shows trait-level QWK on the 20 evaluated CFMS-34 traits.
GRPO-aligned models generally outperform their SFT counterparts, with the Qwen3-based variant achieving the highest average trait QWK and consistent gains on most dimensions.
This indicates that reward-based alignment improves fine-grained rubric judgments in addition to holistic scoring.

Table~\ref{tab:asap_trait_results} reports trait-level QWK on ASAP++ with Qwen2.5.
PPO and GRPO improve most traits over the neural baselines, and prior guidance helps further; HiFTS attains the best mean QWK, indicating that the alignment strategy also generalizes to English multi-dimensional scoring.

\subsection{Feedback Assessment}

Beyond scoring accuracy, Table~\ref{tab:main_results} shows that GRPO-aligned variants are preferred over their SFT counterparts in WinRate, especially on CFMS-34.
This suggests that GRPO improves not only score prediction but also the coherence, rubric alignment, and logical consistency of feedback.
The stronger WinRate of the Qwen3-based model further indicates that stronger base reasoning ability benefits from the proposed reward design.

Because pairwise WinRate relies on an LLM judge, we further report a rule-based grounding metric that checks whether the generated feedback is tied to rubric traits and essay evidence, calculated by:
$0.5\cdot\mathrm{Coverage}+0.3\cdot\mathrm{Trait\text{-}level~Reasoning}+0.2\cdot\mathrm{Reasoning~Density}$.
Coverage is the fraction of scoring traits mentioned in the feedback; trait-level reasoning checks whether each mentioned trait is supported by essay evidence and an evaluative judgment; reasoning density is the fraction of sentences that combine description, explanation, and evaluation. Table~\ref{tab:grounding} shows that reward-aligned variants obtain higher grounding scores than SFT, and full HiFTS performs best.

\begin{table}[ht]
    \small
    \centering
    \setlength{\tabcolsep}{8pt}
    \begin{tabular}{l|c}
        \toprule
        \textbf{Method} & \textbf{Grounding} \\
        \midrule
        HiFTS-SFT & 0.60 \\
        HiFTS-PPO w/o prior & 0.72 \\
        HiFTS-PPO & 0.74 \\
        HiFTS w/o prior & 0.74 \\
        HiFTS & \textbf{0.78} \\
        \bottomrule
    \end{tabular}
    \caption{Rule-based grounding scores of HiFTS-generated feedback on CFMS-34 with Qwen2.5.}
    \label{tab:grounding}
\end{table}

Table~\ref{tab:cot_compare} compares DeepSeek-Reasoner's flat feedback with HiFTS's hierarchical CoT on the same essay. While the flat rationale is relevant, the highlighted annotations indicate that it shifts abruptly across dimensions and lacks clear transitions. In contrast, HiFTS follows a global-to-local process, from overall understanding to rubric dimensions and sub-traits, with highlighted connective expressions linking different dimensions, yielding more cohesive, inspectable, and rubric-aligned feedback.

\subsection{Ablation Study}

Table~\ref{tab:ab_results} analyzes reward-based alignment and prior guidance.
The BERT prior performs moderately, indicating that it is insufficient as a standalone scorer.
However, during LLM inference, it provides coarse quality signals that stabilize hierarchical reasoning, serving as guidance rather than final prediction.

\begin{table}[ht]
    \small
    \centering
    \setlength{\tabcolsep}{6pt}
    \setlength{\aboverulesep}{0.3ex}
    \setlength{\belowrulesep}{0.3ex}
    \begin{tabular}{l|ccc}
        \toprule
        \textbf{Method} & \textbf{Overall} & \textbf{Traits} & \textbf{WinRate} \\
        \midrule
        BERT prior & 0.489 & - & - \\
        \midrule
        Qwen zero-shot & 0.235 & 0.154 & - \\
        Qwen sft w/o CoT & 0.432 & 0.193 & - \\
        HiFTS-SFT & 0.528 & 0.387 & Ref. \\
        HiFTS-PPO w/o prior & 0.572 & 0.417 & 0.603 \\
        HiFTS-PPO & 0.647 & 0.407 & 0.651 \\
        \midrule
        HiFTS w/o prior & 0.579 & 0.420 & 0.621 \\
        HiFTS w/ rand. prior & 0.502 & 0.298 & - \\
        HiFTS w/ const. prior & 0.411 & 0.252 & - \\
        HiFTS & \textbf{0.656} & \textbf{0.439} & \textbf{0.673} \\
        \bottomrule
    \end{tabular}
    \caption{Ablation results on CFMS-34 with Qwen2.5. WinRate is computed against SFT feedback.}
    \label{tab:ab_results}
\end{table}

Comparing Qwen sft w/o CoT with HiFTS-SFT shows that hierarchical feedback-to-score generation improves scoring even without RL or the BERT prior.
Both prior-free PPO and GRPO variants outperform HiFTS-SFT in Overall QWK, confirming the benefit of reward-based alignment.
The GRPO variant also yields a higher WinRate than PPO, indicating better feedback quality.
Adding the prior further improves Overall QWK and WinRate, with HiFTS achieving the best results.
By contrast, random and constant priors degrade performance, showing that guidance must be essay-specific and meaningful.

\begin{figure}[ht]
    \centering
    \includegraphics[width=\linewidth]{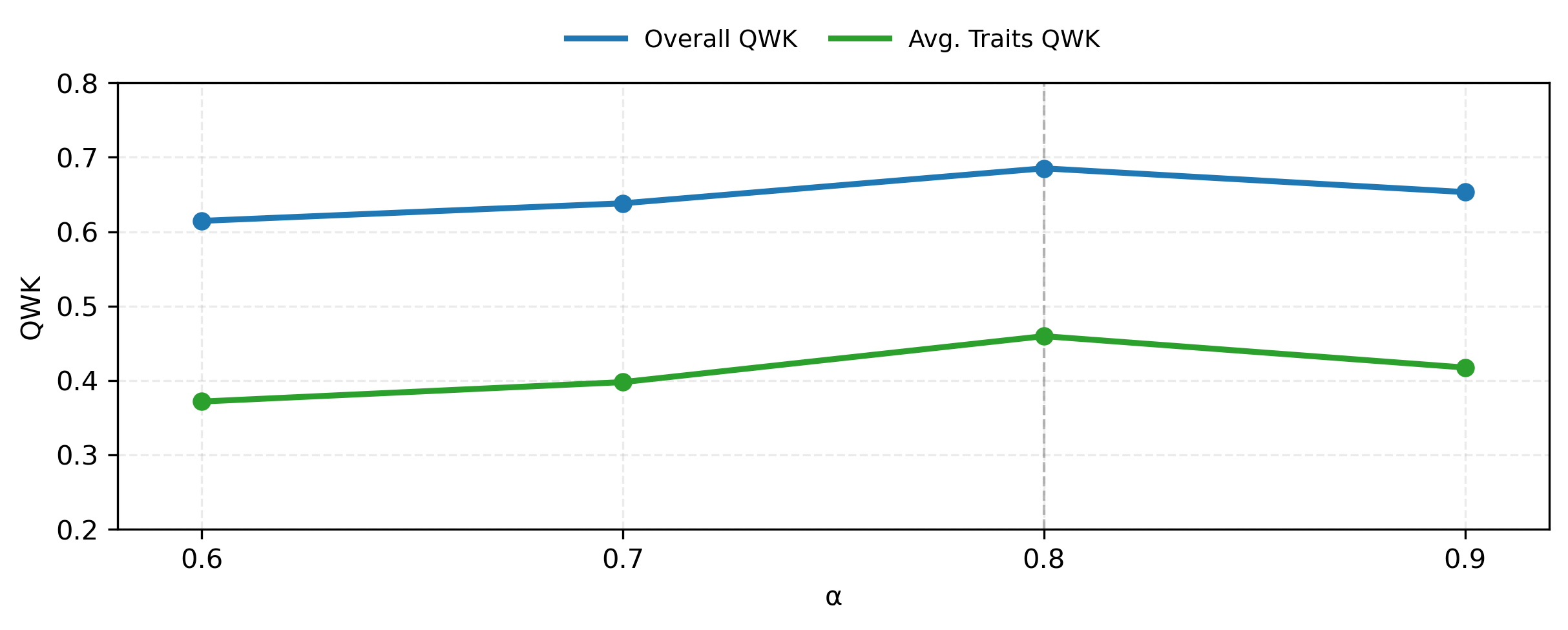}
    \caption{Effect of $\alpha$ on holistic and average trait QWK on the CFMS-34 development set.}
    \label{fig:alpha_ab}
\end{figure}

\subsection{Hyperparameter Analysis}
\label{sec:hyperparameter_analysis}

Figure~\ref{fig:alpha_ab} analyzes the effect of the reward trade-off coefficient $\alpha$.
Smaller values place more emphasis on feedback alignment, while larger values focus more on score-oriented optimization.
We select $\alpha$ on the CFMS-34 development set based on holistic and average trait QWK, which reflect the final agreement of the feedback-to-score process.
Among the tested values, $\alpha=0.8$ achieves the best trade-off and is used in the main experiments.

Table~\ref{tab:window_ab} analyzes the sliding-window size $h$ in Eq.~(4), which estimates holistic QWK over the most recent $h$ generations.
We vary only $h$ on CFMS-34 with HiFTS and Qwen2.5, keeping all other settings fixed.
Smaller windows yield less stable rewards, since QWK is a set-level metric.
$h=128$ performs slightly best, but $h=64$ is close; we use $h=64$ in the main experiments.

\begin{table}[ht]
    \small
    \centering
    \setlength{\tabcolsep}{8pt}
    \begin{tabular}{l|cc}
        \toprule
        \textbf{Window size} & \textbf{Overall} & \textbf{Traits} \\
        \midrule
        16 & 0.612 & 0.422 \\
        32 & 0.641 & 0.432 \\
        64 & 0.656 & 0.439 \\
        128 & 0.659 & 0.443 \\
        \bottomrule
    \end{tabular}
    \caption{Ablation of the sliding-window size $h$ for the holistic QWK reward on CFMS-34 with Qwen2.5.}
    \label{tab:window_ab}
\end{table}

\section{Conclusion}

In this work, we introduced CFMS-34, an expert-annotated Chinese multi-trait AES dataset with dual ratings, holistic scores, and 34 fine-grained rubric-based traits, providing a challenging benchmark for interpretable writing assessment beyond English settings. 
We also proposed HiFTS, a unified autoregressive framework that first generates global-to-local rubric-grounded feedback and then derives holistic and trait-level scores.
HiFTS is trained to learn teacher-generated hierarchical feedback and further aligned with GRPO under a composite reward, while a lightweight BERT-based global prior anchors generation at inference time to improve scoring stability.

Experiments on CFMS-34 and ASAP++ show that HiFTS improves holistic and average trait QWK, reduces MSE on CFMS-34, and produces more coherent rubric-aligned feedback.
Ablation results further verify the benefits of GRPO alignment and prior-guided decoding, suggesting that feedback-to-score reasoning is a promising direction for interpretable multi-trait essay scoring.

\section*{Limitations}

This work has two main limitations. First, HiFTS adopts a predefined global-to-local autoregressive reasoning order for generating hierarchical feedback and scores. Although this structure is effective in our experiments and aligns with rubric-based assessment, exploring alternative generation orders or more flexible feedback organizations may bring further improvements. Second, our evaluation is based on CFMS-34 and ASAP++, covering Chinese and English multi-trait AES settings. To better assess the scalability and educational impact of HiFTS, future studies are needed on more writing genres, learner populations, and real classroom interactions, where students and educators can directly engage with the generated rubric-grounded feedback.

\section*{Ethics Statement}

\paragraph{Use of Scientific Artifacts}
We use ASAP++ as a publicly available AES benchmark, consistent with its intended research use, and use pretrained models and API-based LLMs under their respective licenses or service terms. The creators of the datasets, models, and baselines used in this work are cited in the corresponding sections.

\paragraph{CFMS-34 Data, Annotators, Privacy, and Intended Use}
CFMS-34 is constructed for research on multi-trait automated essay scoring and rubric-grounded feedback generation. The essays were collected from routine educational assessment settings with authorization from the responsible educational institution, and appropriate consent and/or guardian consent procedures were followed according to institutional requirements. The expert raters were Chinese language education specialists from our collaborating education team rather than crowdworkers; they annotated essays according to the predefined rubric described in Section~\ref{sec:cfms34} and Table~\ref{tab:trait_definitions}, and their participation followed the corresponding institutional or collaboration arrangements. All essays were anonymized before annotation, modeling, and release preparation, including removal of direct personal identifiers and screening for potentially identifying content. We release only anonymized essay texts and score annotations under research-use terms, and do not release raw examination records or handwriting images that may contain additional personal information. The dataset and system are intended for research and educational support, not as the sole basis for high-stakes educational decisions without human oversight.

\paragraph{AI Assistants in Writing}
AI assistants were used solely to improve the clarity and coherence of the writing. The ideas and experiments presented in this paper are original to the authors.

\paragraph{Acknowledgements}
This work is supported by the National Natural Science Foundation of China
(62677001), and the Fundamental Research Funds for the Central Universities, Peking University.
\bibliography{custom}

\appendix

\section{Selection of 20 Core Traits}
\label{sec:appendix_traits20}

Although the original CFMS-34 annotations contain 34 fine-grained traits (Table~\ref{tab:trait_definitions}), we select the 20 traits that exhibit the highest correlation with the overall score as the core reasoning subset for modeling and evaluation.

This design is motivated by two considerations:
(1) in real-world grading scenarios, holistic scores are typically dominated by several key traits, while certain traits contribute marginally and may introduce noise;
(2) incorporating too many traits substantially increases the length of the reasoning chain and the complexity of structural constraints, thereby reducing stability in rubric adherence.

Specifically, we compute the Pearson correlation coefficient between each trait score and the overall score on the training set, and select the top 20 traits with the highest correlations.

The selected traits are listed as follows:

\begin{lstlisting}
C02: Able to select concrete and sufficient supporting materials
C04: Able to select and organize materials around the central theme
C05: Able to produce writing with originality and unique insights (depth of thought)
C06: Able to approach the topic from multiple perspectives and select appropriate materials
S02: Able to clearly present preceding and subsequent content
S03: Able to use transitional words and sentences to connect adjacent paragraphs
S06: Each paragraph is closely connected to the main theme
S07: Each sentence is closely connected to the topic sentence of the paragraph
S08: Paragraph-to-paragraph relationships are coherent and tightly connected
S09: Able to elaborate on key parts of the essay in detail
S10: Able to summarize or briefly address transitional content
S11: Able to craft an appropriate introduction
S12: Able to provide a concise and effective conclusion
E01: Sentences are fluent and semantically complete
E02: Word choice is appropriate, vivid, and expressive
E05: Comprehensive use of multiple rhetorical devices
E06: Comprehensive use of multiple modes of expression
E07: Able to employ diverse expressive techniques
E11: Attention to details of characters and events
E12: Observation is structured, prioritized, and sequential
\end{lstlisting}

\section{Per-trait Score Statistics and Inter-rater Agreement}
\label{app:trait_irr}

Table~\ref{tab:trait_irr} reports per-trait score statistics and dual-rater agreement on all 951 CFMS-34 essays.
Mean and standard deviation are computed over the pooled 0--5 ratings of both experts.
Exact is the proportion of identical ratings, Within-1 is agreement within one point, and QWK is quadratic weighted kappa.
The 20 traits used in the main experiments (Appendix~\ref{sec:appendix_traits20}) are marked with $\dagger$.

\begin{table*}[t]
\centering
\small
\setlength{\tabcolsep}{4pt}
\begin{tabular}[t]{lccccc}
\toprule
\textbf{Trait} & \textbf{Mean} & \textbf{Std} & \textbf{Exact} & \textbf{Within-1} & \textbf{QWK} \\
\midrule
Overall & 3.64 & 0.74 & 0.59 & 0.97 & 0.57 \\
\midrule
C01 & 4.08 & 0.86 & 0.40 & 0.87 & 0.25 \\
C02$^{\dagger}$ & 3.96 & 0.84 & 0.43 & 0.87 & 0.32 \\
C03 & 3.63 & 0.81 & 0.45 & 0.90 & 0.34 \\
C04$^{\dagger}$ & 3.81 & 0.79 & 0.50 & 0.89 & 0.33 \\
C05$^{\dagger}$ & 3.23 & 0.86 & 0.49 & 0.93 & 0.52 \\
C06$^{\dagger}$ & 3.59 & 0.79 & 0.52 & 0.93 & 0.41 \\
\midrule
S01 & 3.56 & 0.89 & 0.50 & 0.90 & 0.50 \\
S02$^{\dagger}$ & 3.42 & 0.80 & 0.52 & 0.96 & 0.52 \\
S03$^{\dagger}$ & 3.41 & 0.82 & 0.52 & 0.95 & 0.54 \\
S04 & 3.70 & 0.83 & 0.49 & 0.92 & 0.45 \\
S05 & 3.71 & 0.91 & 0.47 & 0.88 & 0.45 \\
S06$^{\dagger}$ & 3.66 & 0.90 & 0.49 & 0.89 & 0.48 \\
S07$^{\dagger}$ & 3.38 & 0.85 & 0.46 & 0.92 & 0.46 \\
S08$^{\dagger}$ & 3.49 & 0.86 & 0.53 & 0.94 & 0.56 \\
S09$^{\dagger}$ & 3.55 & 0.89 & 0.46 & 0.89 & 0.43 \\
S10$^{\dagger}$ & 3.29 & 0.94 & 0.40 & 0.84 & 0.32 \\
S11$^{\dagger}$ & 3.40 & 0.94 & 0.45 & 0.88 & 0.45 \\
S12$^{\dagger}$ & 3.43 & 0.90 & 0.47 & 0.90 & 0.46 \\
\bottomrule
\end{tabular}
\hspace{1.2em}
\begin{tabular}[t]{lccccc}
\toprule
\textbf{Trait} & \textbf{Mean} & \textbf{Std} & \textbf{Exact} & \textbf{Within-1} & \textbf{QWK} \\
\midrule
E01$^{\dagger}$ & 3.88 & 0.84 & 0.48 & 0.90 & 0.42 \\
E02$^{\dagger}$ & 3.36 & 0.87 & 0.52 & 0.92 & 0.50 \\
E03 & 3.44 & 0.83 & 0.53 & 0.92 & 0.50 \\
E04 & 3.68 & 0.81 & 0.50 & 0.91 & 0.42 \\
E05$^{\dagger}$ & 2.94 & 0.82 & 0.51 & 0.93 & 0.41 \\
E06$^{\dagger}$ & 3.02 & 0.82 & 0.47 & 0.91 & 0.35 \\
E07$^{\dagger}$ & 2.92 & 0.79 & 0.49 & 0.92 & 0.36 \\
E08 & 3.67 & 0.81 & 0.55 & 0.91 & 0.46 \\
E09 & 2.56 & 0.78 & 0.69 & 0.96 & 0.64 \\
E10 & 3.58 & 0.79 & 0.52 & 0.90 & 0.38 \\
E11$^{\dagger}$ & 3.53 & 0.84 & 0.54 & 0.90 & 0.45 \\
E12$^{\dagger}$ & 3.69 & 0.83 & 0.47 & 0.87 & 0.32 \\
\midrule
Cv01 & 4.21 & 0.85 & 0.31 & 0.79 & $-$0.01 \\
Cv02 & 4.22 & 0.82 & 0.36 & 0.87 & 0.17 \\
Cv03 & 3.94 & 0.88 & 0.52 & 0.90 & 0.49 \\
Cv04 & 3.55 & 0.93 & 0.36 & 0.77 & 0.12 \\
\bottomrule
\end{tabular}
\caption{Per-trait score statistics and inter-rater agreement on CFMS-34 (951 essays). $\dagger$ denotes the 20-trait subset used in the main experiments.}
\label{tab:trait_irr}
\end{table*}

\section{Example of CoT Generation Prompt Design}
\label{sec:appendix_prompt}

The following prompt is used to guide the teacher model to generate rubric-grounded hierarchical feedback.
The scoring rubric follows the core trait definitions in Appendix~\ref{sec:appendix_traits20}.

\begin{lstlisting}[breaklines=true, columns=fullflexible]
You are an experienced Chinese language teacher with many years of essay grading experience.
Please read the student essay and the given trait scores (0 to 10), and generate a hierarchical evaluative comment organized into three levels:

[Hierarchical Structure]
1. Content: Analyze indicators related to material selection, theme alignment, originality, depth of thought, and multi-perspective thinking.
2. Structure: Analyze paragraph organization, transitions, logical sequencing, coherence, and the handling of the introduction and conclusion.
3. Expression: Analyze sentence fluency, word choice, rhetorical devices, modes of expression, expressive techniques, descriptive details, and observation order.

[Core Requirements]
- Scope of Analysis: You must analyze only the indicators explicitly provided with scores in the input, such as C02, S02, or E01. The use of generalized category codes, such as Cxx, Sxx, or Exx, is strictly prohibited.
- Indicator Anchoring: Whenever a specific evaluation point is discussed, append the corresponding indicator code in parentheses at the end of the sentence, without spaces, e.g., (C02).
- Structured Output: The analysis must strictly follow the order Content -> Structure -> Expression.
- Style Constraint: Encouraging or congratulatory language, such as "Keep up the good work", is strictly prohibited.
- Formatting Constraint: Use natural language for analysis without bold formatting, bullet points, or special symbols in the final response.

[Scoring Rubric Reference]: Appendix A.
Use only the provided core trait definitions and their corresponding scores.
\end{lstlisting}

\section{Example of Scoring System Prompt Design}
\label{sec:appendix_system_prompt}

\begin{lstlisting}
You are a primary school Chinese language teacher with many years of grading experience. Your task is to read the student essay and strictly follow the provided scoring rubric and evaluation procedure to generate a hierarchical evaluative comment.

[Scoring Rubric Reference]: Appendix A.

[Evaluation Procedure]
1. Analysis: First, provide a detailed analysis of the essay across all fine-grained dimensions according to the scoring rubric. The analysis must follow the hierarchical logic of each dimension.
2. Sub-scores: Second, provide the specific scores (0~10) for all 20 fine-grained dimensions.
3. Overall Score: Finally, provide the total score of the essay.

[Output Format]
The output must contain exactly the following three top-level tags:
1. [Analysis]: Provide detailed chain-of-thought analysis and hierarchical evaluative comments.
2. [Trait Scores]: Provide all fine-grained scores line by line (e.g., C02:7.0).
3. [Overall Score]: Provide the final overall score of the essay.

Strict adherence to the [Output Format] requirements is mandatory.
\end{lstlisting}

\end{document}